\documentclass[12pt]{article}

\usepackage[export]{adjustbox}
\usepackage[ruled,vlined,linesnumbered]{algorithm2e}
\usepackage{listings} 
\RestyleAlgo{boxed}

\usepackage{algpseudocode}
\usepackage{amscd,amsfonts,amsopn,amssymb,amstext}
\usepackage{appendix}
\usepackage{bbm,bm}
\usepackage{natbib}
\usepackage{url}
\usepackage{fullpage}
\usepackage{graphics,psfrag}
\usepackage{latexsym}
\usepackage{lscape}
\usepackage{rotating}
\usepackage{srcltx}
\usepackage{threeparttable}
\usepackage{times}
\usepackage{verbatim}
\usepackage{bm,bbm}
\usepackage{ulem}
\usepackage{braket}
\usepackage{mathrsfs}
\usepackage{mathtools}
\usepackage{smartdiagram}
\usesmartdiagramlibrary{additions}
\usetikzlibrary{arrows}
\usetikzlibrary{shapes}
\usepackage{lscape}
\usepackage{array,longtable}
\usepackage{pdflscape} 

\usepackage{authblk}
\usepackage{subfig}
\usepackage{tikz}

\usepackage{booktabs} 
\usepackage{siunitx} 

\usepackage{comment}

\begin{document}
	
	\def\spacingset#1{\renewcommand{\baselinestretch}%
		{#1}\small\normalsize} \spacingset{1}

	
	\noindent {\textbf{\Large A Novel Hybrid Approach Using an Attention-Based\\
			Transformer $+$ GRU Model for Predicting Cryptocurrency Prices}}
	
	\vskip 8mm
	\noindent Esam Mahdi$^{1}$; Martin-Barreiro, C.; Cabezas, X.
	
	\noindent $^{1}$ Corresponding author: School of Mathematics and Statistics, Carleton University, Canada
	
	\noindent EsamMahdi@cunet.carleton.ca

	\bigskip
	\begin{abstract}
		In this article, we introduce a novel deep learning hybrid model that integrates attention Transformer and Gated Recurrent Unit (GRU) architectures to improve the accuracy of cryptocurrency price predictions. 
		By combining the Transformer's strength in capturing long-range patterns with the GRU's ability to model short-term and sequential trends, the hybrid model provides a well-rounded approach to time series forecasting.
		We apply the model to predict the daily closing prices of Bitcoin and Ethereum based on historical data that include past prices, trading volumes, and the Fear and Greed index. 
		We evaluate the performance of our proposed model by comparing it with four other machine learning models: two are non-sequential feedforward models: Radial Basis Function Network (RBFN) and General Regression Neural Network (GRNN), and two are bidirectional sequential memory-based models: Bidirectional Long-Short-Term Memory (BiLSTM) and Bidirectional Gated Recurrent Unit (BiGRU).
		The performance of the model is assessed using several metrics, including Mean Squared Error (MSE), Root Mean Squared Error (RMSE), Mean Absolute Error (MAE), and Mean Absolute Percentage Error (MAPE), along with statistical validation through the nonparametric Friedman test followed by a post hoc Wilcoxon signed rank test.
		The results demonstrate that our hybrid model consistently achieves superior accuracy, highlighting its effectiveness for financial prediction tasks. 
		These findings provide valuable insights for improving real-time decision making in cryptocurrency markets and support the growing use of hybrid deep learning models in financial analytics.
	\end{abstract}
	
	\noindent%
	\vfill

	\section{Introduction}
	
	The cryptocurrency market has grown exponentially, with Bitcoin and Ethereum standing out as the top two dominant digital assets.
	Both accounts for the majority of trading activity and significantly impact global financial trends.
	One of the biggest challenges in working with cryptocurrencies is figuring out how to develop accurate models that can predict their highly volatile prices.
	Numerous studies have showed that machine learning (ML) models are highly effective in predicting cryptocurrency prices, as they can capture complex nonlinear patterns in financial time series \cite{Razi2005,Slepaczuk2018,Chen2020,Mahdi2021,Akyildirim2021,Makala2021,Jaquart2022,Mahdi2022,Qureshi2024}. 
	Among these, artificial neural network (ANN)-based machine learning algorithms, such as radial basis function network (RBFN), general regression neural network (GRNN), gated recurrent unit (GRU), and long-short-term memory (LSTM), have been widely applied in this field.
	
	\citet{Broomhead19881,Broomhead19882} introduced the radial basis function networks (RBFNs) to model complex relationships between features and target variables and to make predictions on new data. 
	RBFNs are a type of simple feedforward neural network that utilizes radial basis functions as their activation functions.
	One of the key benefits of using radial basis functions in RBFNs is their ability to smooth out non-stationary time series data while effectively still modeling the underlying trends.
	\citet{Alahmari2020} compared using the linear, polynomial, and radial basis function (RFB) kernels in the support vector regression for predicting the prices of Bitcoin, XRP and Ethereum and showed that the RFB outperforms the other kernel methods in terms of accuracy and effectiveness.
	Recently, \citet{Casillo2022} showed that the RRBFn models are effective in predicting Bitcoin prices from the analysis of online discussion sentiments.
	More recently, \citet{Zhang2025} demonstrated that combining a radial basis function (RBF) with a battle royale optimizer (BRO) significantly enhances the accuracy of stock price predictions, including those for cryptocurrencies. 
	The results of his study showed that this approach outperformed other machine learning models like LSTM, BiLSTM-XGBoost, and CatBoost.
	
	The general regression neural network (GRNN) is another type of feedforward neural network was first proposed by \citet{Specht1991} designed for efficient use in regression tasks. 
	One of the key strengths of GRNN is its ability to model complex nonlinear relationships without the need for iterative training. 
	This has made it a powerful tool for applications in regression, prediction, and classification.
	Despite its advantages, GRNN's application in cryptocurrency prediction is still relatively not widely explored \cite{Martinez2019,Martinez2022}. 
	In this article, we try to fill in the gap by assessing the performance of GRNN in predicting the prices of Bitcoin and Ethereum.
	
	Recurrent neural networks (RNNs) are designed for handling sequential data because, unlike feedforward neural networks, they have feedback loops that enable them to retain information from previous inputs using a method called backpropagation through time (BPTT). 
	However, when training on long sequences, the residual error gradients that needs to be propagated back diminishes exponentially due to repeated multiplication of small weight across time steps. 
	This makes it difficult for the network to learn long-term dependencies.
	To address such a problem and to improve the modeling of long-term dependencies in sequential data, \citet{Hochreiter1997} proposed the long-short-term memory (LSTM) network model. 
	LSTM model has been widely adopted in literature, as it has proven to be highly effective in forecasting cryptocurrency prices, outperforming other machine learning models \cite{McNally2018,Liu2020,Zoumpekas2020}.
	\citet{Lahmiri2019} showed that the Bitcoin, Digital Cash, and Ripple price predictability of LSTM is significantly higher when compared to that of GRNN.
	\citet{Ji2019} conducted a comparative study comparing the LSTM network model with deep learning network, convolutional neural network, and deep residual network models for predicting the Bitcoin prices and showed that the LSTM slightly outperformed the other models for regression problems, whereas for classification (up and down) problems, deep learning network works the best.
	\citet{Uras2020} used LSTM to predict the daily closing price series of Bitcoin, Litecoin and Ethereum cryptocurrencies, based on the prices and volumes of prior days and showed that the LSTM outperformed the traditional linear regression models.
	\citet{Lahmiri2021} conducted a study comparing the performance of LSTM and GRNN models in predicting Bitcoin, Digital Cash, and Ripple prices. Their findings indicated that the LSTM model outperformed GRNN in terms of prediction accuracy.
	
	The gated recurrent unit (GRU) neural network model, introduced by \citet{Cho2014}, was designed to handle long-term dependencies in sequential data, similar to the LSTM model, but with fewer parameters, making GRU computationally less expensive and faster than LSTM \cite{Jianwei2019}.
	\citet{Dutta2020} used GRU to predict the prices of cryptocurrencies and demonstrated that the GRU outperformed the LSTM networks.
	\citet{Tanwar2021} employed GRU and LSTM to predict the price of Litecoin and Zcash cryptocurrencies, taking into account the inter-dependency of the parent coin. 
	Their findings showed that these models forecasted the prices with high accuracy compared to other machine learning models.
	\citet{Ye2022} proposed a method combining LSTM and GRU to predict the prices of Bitcoin using the historical transaction data, sentiment trends of Twitter, and technical indicators, and showed that their method can better assist investors in making the right investment decision.
	\citet{Patra12023} developed a multi-layer GRU network model with multiple features to predict the prices of Bitcoin, Ethereum, and Dogecoin, and demonstrated that  their model provided a better performance compared with the LSTM and GRU models that  used a single feature.
	
	Several studies have shown that bidirectional long-short-term memory (BiLSTM) and bidirectional gated recurrent unit (BiGRU) enhance the accuracy of financial time series predictions. Unlike traditional LSTM and GRU, these bidirectional models process data in both forward and backward directions allowing to capture dependencies from past and future time steps. 
	For example, \citet{Hansun2022} compared three popular deep learning architectures, LSTM, bidirectional LSTM, and GRU, for predicting the prices of five cryptocurrencies, Bitcoin, Ethereum, Cardano, Tether, and Binance Coin using various prediction models. 
	Their findings indicated that BiLSTM and GRU performed just as well as LSTM, offering robust and accurate predictions of cryptocurrency prices.
	\citet{Ferdiansyah2023} showed that combining GRU and BiLSTM in a hybrid model increased prediction accuracy for Bitcoin, Ethereum, Ripple, and Binance.
	
	In this article, we propose a novel hybrid transformer $+$ GRU model and compare its performance with two non-sequential feedforward models (GRNN and RBFN) and two bidirectional sequential memory-based models (BiGRU and BiLSTM) for forecasting cryptocurrency prices across two distinct modeling scenarios. 
	The first scenario aims to predict Bitcoin prices using historical Bitcoin price data, its trading volume, and crypto fear and greed index (FGI). 
	The second scenario aims to predict Ethereum prices using historical price data for both Bitcoin and Ethereum, along with Ethereum's trading volume and the Fear and Greed Index (FGI).
	The Transformer model, introduced by \citet{Vaswani2017}, revolutionized natural language processing (NLP) by using self-attention and feed-forward networks.
	Since then, transformers have been adapted and extended in numerous ways and became the foundation for AI systems like ChatGPT, DeepSeek, and others \cite{Devlin2018, Dosovitskiy2020}.
	{\color{black}The application of transformer neural networks to financial time series data is a promising area of research \cite{Zhou2021,Grigsby2021,Lezmi2023,Wen2023}. 
		For example, \citet{Li2020} successfully applied Transformers to forecast both synthetic and real-world time series data. Their work demonstrated that Transformers excel at capturing long-term dependencies, an area where LSTMs often struggle. 
		\citet{Zerveas2021} demonstrated that the transformer-based approach for unsupervised pre-training on multivariate time series outperforms traditional methods for both regression and classification tasks, even with smaller datasets.
		\citet{Wu2021} developed an Autoformer model based on decomposing Transformers with autocorrelation mechanism to better handle long-term forecasting and seasonality in complex time series. They showed that their model outperformed traditional approaches, delivering more accurate results.
		Recent work by \citet{Castangia2023} highlights how effective Transformer models are for flood forecasting, outperforming  traditional recurrent neural networks (RNNs).
		More recently, \citet{Nayak2024} proposed a user-friendly \texttt{Python} code that implements transformer architecture specifically for time series forecasting data.}
	
	To the best of our knowledge, this is the first study to introduce a deep learning model that combines a parallel self-attention-based Transformer architecture with a sequential memory-based GRU model.   
	{\color{black}This unique combination enables our model to effectively capture both long-term dependencies through the Transformer's self-attention mechanism and short-term temporal patterns using the GRU’s gated memory. 
		This makes the model well-suited to capturing market dynamics across different time scales, especially during volatile periods.}
	The rest of this article is organized as follows:
	Section \ref{sec:PredictionModels} defines the prediction models used to predict Bitcoin and Ethereum prices. 
	Here, we define four neural network models that we need to compare with our proposed model.
	Two of these four models are classified as feedforward neural networks and the other two are classified as memory-based models.
	Section \ref{sec:Transformer} introduces our new deep learning approach for cryptocurrency price prediction.
	Section \ref{sec:Analysis} presents the data and results, showcasing how all five models perform in this comparative study.
	Finally, Section \ref{sec:Conclusions} summarizes our findings and draws some conclusions and recommendations.
	
	\section{Prediction Models}\label{sec:PredictionModels}
	
	We consider modeling the cryptocurrency prices as a function of past feature values with a one-time-step lag:
	\begin{equation}\label{eq:model}
		y_t = f\left(\mathbf{x}_{t-1}\right)+\varepsilon_{t},
	\end{equation}
	where $y_t$ is the target variable represents the price of Bitcoin or Ethereum at time $t$, $f\left(\mathbf{x}_{t-1}\right)$ is a function of the features $\mathbf{x}_{t-1}$ at time $t-1$ that need to be approximated, and $\varepsilon_t$ is the error term represents the difference between the predicted price and actual price at time $t$.
	Her, we consider $\mathbf{x}_{t-1}$ is a three dimensional vector represents the prices of Bitcoin/Ethereum, $P_{t-1}$, in US dollars, fear and greed index (FGI), $F_{t-1}$, and exchange trading volume, $V_{t-1}$, (USD) of Bitcoin/Ethereum at time $t-1$.
	
	We use the following four different network along with our proposed model to forecast the prices Bitcoin and Ethereum given in Equation \ref{eq:model}. 
	
	
	\subsection{Radial Basis Function Network (RBFN)}\label{sec:RBFN}
	
	Radial basis function network (RBFN) is a type of feedforward neural network that uses radial basis functions as activation functions.
	Typically, it consists of three layers as seen in Figure \ref{fig:RBFN}: the Input Layer, which receives the input data; the Hidden Layer, which  applies radial basis functions, such as Gaussian activation (kernel) functions, to transform the input data into a higher-dimensional space; and the Output Layer, which produces the final output, often as a linear combination of the hidden layer outputs.
	\usetikzlibrary{positioning}
	\begin{figure}[htb!]
		\centering
		\begin{tikzpicture}[node distance=1.3cm]
			
			\node[draw, circle, xshift=1.5cm, yshift=0.75cm] (hidden1) {\(\phi_1(\cdot)\)};
			\node[draw, circle, below of=hidden1] (hidden2) {\(\phi_2(\cdot)\)};
			\node[draw, circle, below of=hidden2] (hidden3) {\(\phi_m(\cdot)\)};
			\node[above of=hidden1, yshift=-0.5cm] (hiddenLabel) {Hidden Layer};
			
			\node[draw, rectangle, left of =hidden2] (input) at (-.55, -.55) {
				\begin{tabular}{c|c|c|c}
					\(X_{1}\) & \(X_{2}\) & \(\cdots\) & \(X_{k}\) \\
				\end{tabular}
			};
			\node[above of=input, yshift=-0.5cm] (inputLabel) {Input Layer};
			
			\node[draw, circle, right of=hidden2, xshift=1.5cm] (output) {\(\Sigma\)};
			\node[above of=output, xshift=0.8cm,yshift=-0.8cm] (outputLabel) {Output Layer};
			
			\draw[->] (input.east) -- ++(0.75,0) |- (hidden1.west);
			\draw[->] (input.east) -- ++(0.75,0) |- (hidden2.west);
			\draw[->] (input.east) -- ++(0.75,0) |- (hidden3.west);
			
			\draw[->] (hidden1.east) -- node[pos=0.4, above] {\(w_1\)} (output.west);
			\draw[->] (hidden2.east) -- node[pos=0.4, above] {\(w_2\)} (output.west);
			\draw[->] (hidden3.east) -- node[pos=0.4, above] {\(w_m\)} (output.west);
			
			\draw[->] (output.east) -- ++(0.75,0) node[right] {\(y = f(\mathbf{x})+w_{0}\)};
		\end{tikzpicture}
		\caption{Architecture of a radial basis function network (RBFN).}	
		\label{fig:RBFN}
	\end{figure}
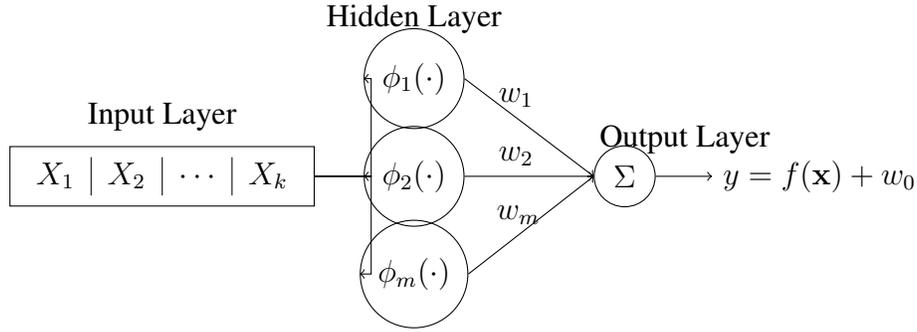
	
	Mathematically, the input can be considered as a vector of $k$ variables $\mathbf{x} = (X_1, X_2,\cdots, X_k)^\top$, each with $n$ observations. 
	The kernel function is used as the activation function in the hidden layer.
	Typically, the most common choice is the Gaussian function:
	\begin{equation}\label{eqn:kernel}
		\varphi\left(\mathbf{x}, \mathbf{c}_i\right)=\exp \left(-\frac{\left\|\mathbf{x}-\mathbf{c}_i\right\|^2}{2 \sigma_i^2}\right)
	\end{equation}
	where $||\mathbf{x} -\mathbf{c}_i ||$ is the Euclidean distance between the input vector $\mathbf{x}$ and the center $\mathbf{c}_i = (c_{i1},c_{i2},\cdots,c_{ik})$, and $\sigma_i$ is the bandwidth or spread parameter.
	The output of the hidden layer is passed to the output layer, where the final output is computed as a linear combination of the hidden layer outputs:
	\begin{equation}\label{eqn:RBFN}
		y=\sum_{i=1}^m w_i \varphi\left(\mathbf{x}, \mathbf{c}_i\right)+w_{0}
	\end{equation}
	where $m$ denotes the number of neurons in the hidden layer, $\mathbf{c}_i$ is the center vector for neuron $i$, $w_i$ is the weight of neuron $i$ in the linear output neuron, $w_{0}$ is the bias term, and $\varphi$ is the activation function (radial basis function) for the $i^{\text{th}}$ RBF neuron. 
	
	\subsection{General Regression Neural Network (GRNN) Model}\label{sec:GRNN}
	
	The general regression neural network (GRNN), introduced by \citet{Specht1991}, is another type of feedforward neural network that is designed for regression tasks and closely related to the radial basis function (RBF) network. 
	Unlike traditional feedforward and deep neural networks, GRNN does not rely on backpropagation or gradient-based optimization. 
	Instead, it operates by minimizing the difference between predicted and actual target values. 
	The model directly memorizes the training data and leverages it for predictions. 
	
	GRNN comprises three layers: the input layer, hidden layer, and output layer.
	The set of $n$ training data $\{\mathbf{x}_i, y_i\}_{i=1}^{n}$ enters the input layer as a set of $k$ features $\mathbf{x}$ and their associated target $y$.
	Then the hidden layer uses a kernel function to compute the similarity between a new input feature $\mathbf{x}$ and each training sample $\mathbf{x}_i$.
	The Gaussian kernel is commonly used:
	\begin{equation}\label{eqn:kernel2}
		K\left(\mathbf{x}, \mathbf{x}_i\right)=\exp \left(-\frac{\left\|\mathbf{x}-\mathbf{x}_i\right\|^2}{2 \sigma_i^2}\right),
	\end{equation}
	where $\mathbf{x}, \mathbf{x}_i$, and $\sigma_i$ are the new input vector, the training sample, and the smoothing parameter respectively.
	Then it gives a set of weights associated with the closeness distance:
	\begin{equation}\label{eqn:weightsKernel}
		w_i = \frac{\exp \left(-\frac{\left\|\mathbf{x}-\mathbf{x}_i\right\|^2}{2 \sigma_i^2}\right)}{\sum_{i=1}^{n}\exp \left(-\frac{\left\|\mathbf{x}-\mathbf{x}_i\right\|^2}{2 \sigma_i^2}\right)}.
	\end{equation}  
	Finally, the output layer computes the prediction value of $y$ as a weighted average of nearby observations:
	\begin{equation}\label{eqn:predGRNN}
		\hat{y} = \sum_{i=1}^{n} w_i y_i
	\end{equation} 
	
	
	\subsection{long-short-term Memory (LSTM) Model}\label{sec:LSTM}
	
\begin{figure}[ht]
		\centering
		\includegraphics*[width=0.7\textwidth,height=0.24\textheight]{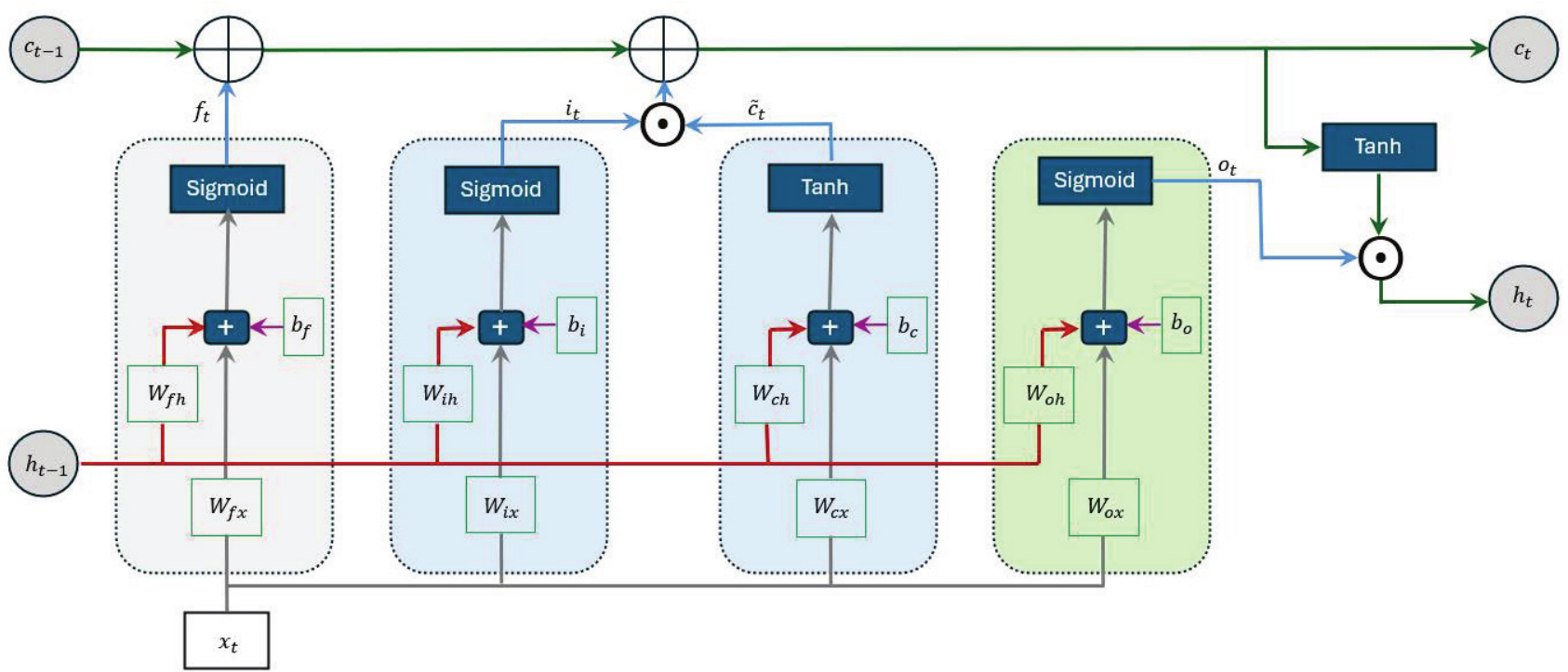}
		\vspace{-0.1cm}
		\caption{Architecture of long-short-term memory (LSTM) network contains four interacting layers.}
		\label{fig:fig1}
	\end{figure}

	Long-short-term memory (LSTM) is a specialized type of recurrent neural network (RNN) developed to solve the vanishing gradient problem that standard RNNs often faced.
	This problem makes it challenging for the model to capture long-term dependencies when dealing with long sequences.
	To overcome this limitation, LSTM uses a memory cell designed to store information over extended periods while discarding irrelevant details. 
	Figure \ref{fig:fig1} illustrates the structure of the LSTM.
	
	The first step of the LSTM network is to decide what information will be discarded from the memory cell state $c_{t-1}$.
	The sigmoid activation function in the forget gate is applied to the current input, $X_t$, and the output from the previous hidden state, $h_{t-1}$, and produces values between 0 and 1 for each element in the cell state. 
	A value of 1 indicates "keep this information", while a value of 0 means "omit this information":
	\begin{equation}\label{eqn:LSTM1}
		\text{Forget gate:}\quad f_t=\sigma\left(W_{fx} X_t+W_{f h} h_{t-1}+b_f\right)
	\end{equation}
	where $W_{fx}$ and $W_{f h}$ represent the weight matrices for the input and recurrent connections, respectively, and $b_f$ denotes the bias vector parameters. 
	
	The next step decides what new information will be stored in the cell state.
	The input gate first applies the sigmoid layer to the input from the previous hidden state and the input from the current state to determine which parts of the information should be updated.   
	Then, a tanh layer generates a new candidate value, $\tilde{c}_t$, that could potentially be added to the cell state:
	\begin{align}\label{eqn:LSTM2}
		\text{Input gate:}\quad i_t & =\sigma\left(W_{i x} x_t+W_{i h} h_{t-1}+b_i\right) \\
		\text{Intermediate cell state:}\quad \widetilde{c_t} & =\tanh\left(W_{c x} x_t+W_{c h} h_{t-1}+b_c\right)
	\end{align}
	where $W$ and $b$ are weight matrices and bias vector parameters.
	Following this, the previous cell state $c_{t-1}$ is updated by a new cell state $c_t$:
	\begin{equation}\label{eqn:LSTM3}
		\text{New cell state:}\quad c_t=f_t \odot c_{t-1}+i_t \odot \tilde{c}_t 
	\end{equation}
	where $\odot$ denotes the Hadamard product (element-wise product).
	
	Lastly, the model determines the final output which is derived from the updated cell state after applying a filtering process: 
	First, a sigmoid layer decides which portions of the cell state will be included in the output. 
	Next, the cell state is passed through a tanh activation function to scale its values between $-1$ and $1$, and the result is multiplied by the output gate to produce the final output.
	\begin{align}\label{eqn:LSTM4}
		\text{Output gate:}\quad o_t &=\sigma\left(W_{o x} x_t+W_{o h} h_{t-1}+b_o\right) \\
		\text{New state:}\quad h_t &=o_t  \odot\tanh\left(c_t\right)
	\end{align}
	
	\subsection{Gated Recurrent Unit (GRU) Model}\label{sec:GRU}
	
	The Gated Recurrent Unit (GRU) has a simpler design than the LSTM network while still being effective at capturing long-term dependencies and handling the vanishing gradient problem we face in using RNN \citet{Cho2014}. 
	It uses two gates: the reset gate and update gate. 
	The reset gate determines how much of the previous hidden state $h_{t-1}$ should be discarded when computing the new candidate hidden state $\tilde{h}_{t-1}$.
	The update gate controls how much of the new candidate hidden state, $\tilde{h}_{t-1}$, should be used to update the current hidden state.
	
	Figure \ref{fig:fig2} illustrates the structure of the GRU, while Equations \ref{eqn:GRU1}$-$\ref{eqn:GRU4} explain its functionality mathematically as follows:
	\begin{figure}[ht]
		\centering
		\includegraphics*[width=0.7\textwidth,height=0.23\textheight]{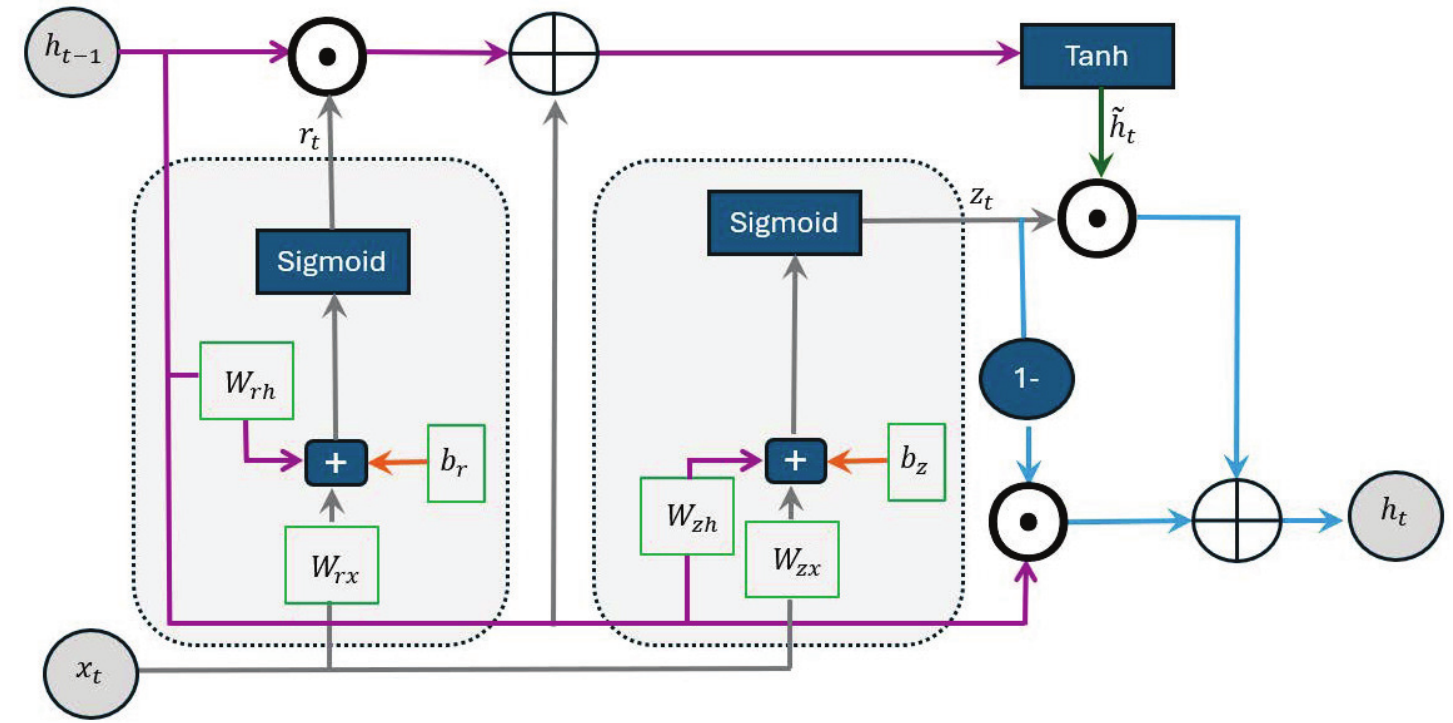}
		\vspace{-0.1cm}
		\caption{Architecture of gated recurrent unit (GRU) network.}
		\label{fig:fig2}
	\end{figure}
	\begin{align}
		\text{Reset gate}:&\quad r_t =\sigma\left(W_{r x} x_t+W_{r h} h_{t-1}+b_r\right)\label{eqn:GRU1}\\		
		\text{Update gate}:&\quad z_t =\sigma\left(W_{z x} x_t+W_{z h} h_{t-1}+b_z\right)\label{eqn:GRU2}\\	
		\text{Cell state}:&\quad \tilde{h}_t =\tanh \left(W_x x_t+W_h\left(r_t \odot h_{t-1}\right)+b\right)\label{eqn:GRU3}\\
		\text{New state}:&\quad h_t  =z_t \odot \tilde{h}_t+\left(1-z_t\right) \odot h_{t-1}\label{eqn:GRU4}
	\end{align}
	where, $x_t$ denotes the input at the current state, $h_{t-1}$ represents the hidden state at the previous state, $r_t$ signifies the output of the reset gate, and $z_t$ corresponds to the output of the update gate.
	The symbol $\odot$ stands for the Hadamard product, which is an element-wise multiplication operation and $\tilde{h}_{t}$ denotes the candidate hidden state. 
	Additionally, $W_x$ and $W_h$ refer to the feedforward and recurrent weight matrices, respectively, while $b$ represents the bias parameters.
	
	As shown in Figure \ref{fig:fig2} and Equations \ref{eqn:GRU1}$-$\ref{eqn:GRU4}, the reset gate applies the sigmoid activation function to a linear combination of the current state, the output from the previous state, and a bias term. 
	This helps determine how much information should be discarded.
	Similarly, the update gate uses the sigmoid function on a different linear combination of the current state, the previous output, and a bias term to decide how much information should be updated.
	The tanh activation function is then used to generate a new candidate value,  $\tilde{h}_t$. 
	This candidate is multiplied (using the Hadamard product) by the output from the update gate. 
	Meanwhile, the difference value of the update state from 1 is multiplied by the previous output (again using the Hadamard product), and  this result is added to the first product to produce the final output.
	
\begin{figure}[ht]
		\centering
		\includegraphics*[width=0.7\textwidth,height=0.23\textheight]{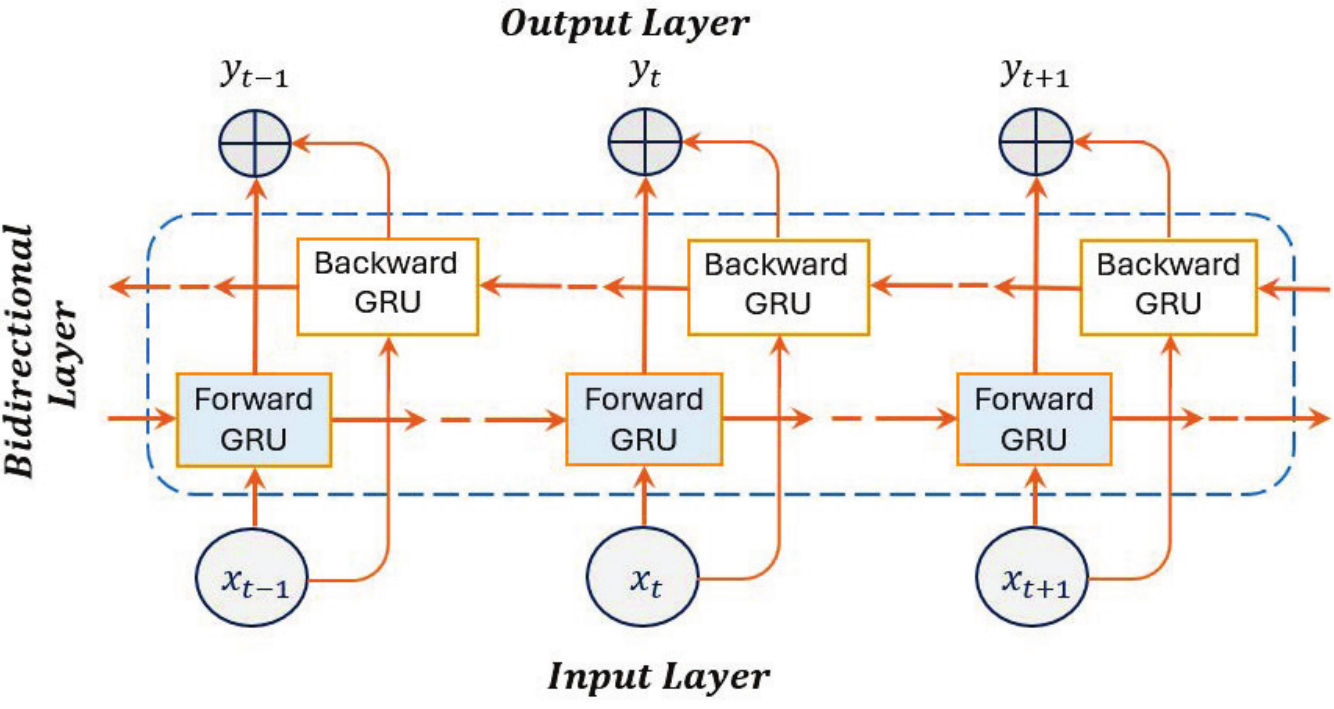}
		\vspace{-0.1cm}
		\caption{Architecture of bidirectional gated recurrent unit (BiGRU).}
		\label{fig:fig3}
	\end{figure}

	Research has demonstrated that processing the input sequences in both forward and backward directions can improved the accuracy of time series predictions. 
	In our analysis, we will consider the two bidirectional models: the bidirectional long-short-term memory (BiLSTM) and the bidirectional gated recurrent unit (BiGRU).
	Figure \ref{fig:fig3} shows the architecture of BiGRU. 
	The BiLSTM follows the same architecture as the BiGRU shown in Figure \ref{fig:fig3}, replacing the word "GRU" by "LSTM".
	The forward pass in BiLSTM and BiGRU processes the sequence from $t=1$ to $t=n$, while the backward pass processes the sequence from $t=n$ to $t=1$.
	This bidirectional architecture allow these models to capture dependencies from past and future time steps, leading to improved prediction performance.  
	
	\section{Hybrid Transformer $+$ GRU Architecture}\label{sec:Transformer}
	
	Transformer neural networks, introduced by \citet{Vaswani2017} in their seminal 2017 paper "Attention is All You Need", have revolutionized the field of natural language processing (NLP). 
	Unlike traditional sequential models like LSTM and GRU, transformers use self-attention to model relationships between all elements in a  parallel sequence simultaneously, rather than processing them step by step.  
	This architecture allows transformers to handle both local and global dependencies without the need for recurrent connections more efficiently.
	The transformers follow an encoder-decoder architecture, where both consist of multiple identical layers.
	The encoder processes the input, and the decoder produces the output.
	By leveraging the strengths of transformer and gated recurrent unit deep learning models, we propose a new hybrid Transformer $+$ GRU model for cryptocurrency price prediction. 
	
\begin{figure}[ht]
		\centering
		\includegraphics*[width=0.7\textwidth,height=0.23\textheight]{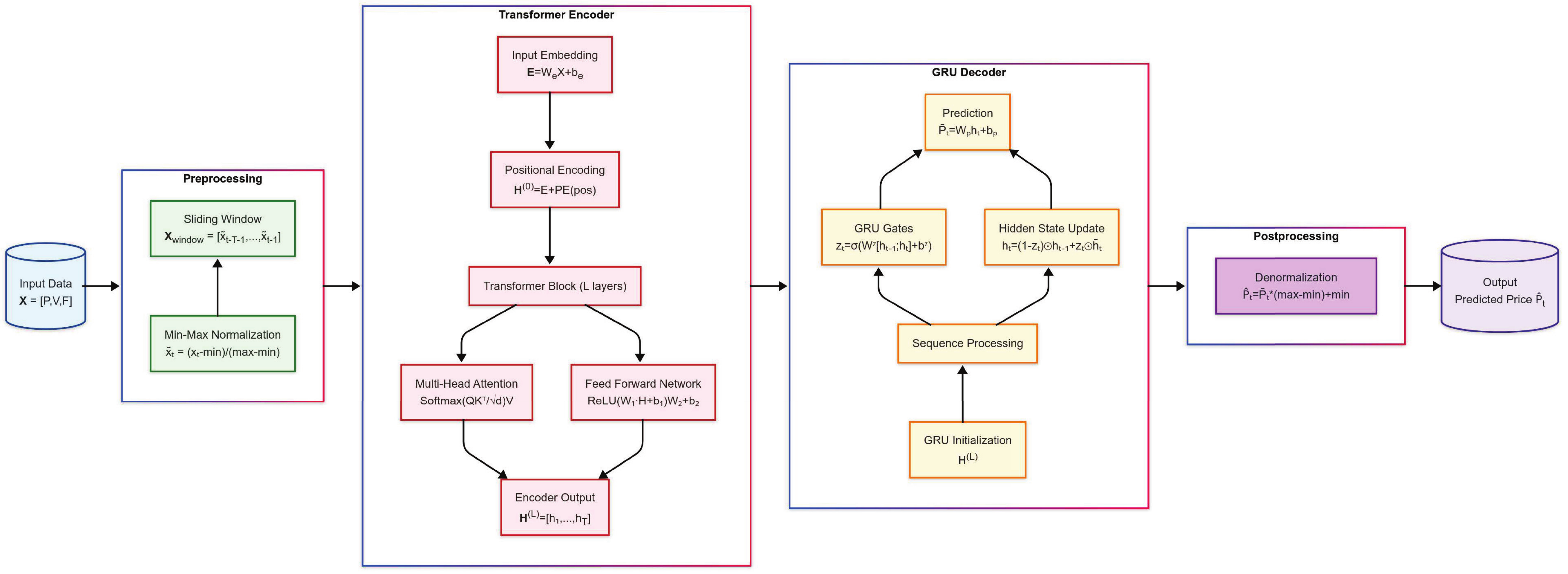}
		\vspace{-0.1cm}
		\caption{Architecture of hybrid Transformer $+$ GRU model.}
		\label{fig:fig4}
	\end{figure}

	{\color{black}\subsection{Step-by-step methodology}\label{sec:methodology}
		Below is a step-by-step algorithm of how the hybrid Transformer + GRU model works:
		\begin{description}
			\item{\it \textbf{1. Input features:}} At each time $i$,
			$$
			x_i=\left[P_i, V_i, F_i\right]\in\mathbb{R}^3,
			$$
			where $P_i, V_i, F_i$ are the price of Bitcoin/Ethereum, exchange trading volume of Bitcoin/Ethereum, and fear and greed index, respectively.
			
			\item{\it \textbf{2. Preprocessing:}}
			\begin{itemize}
				\item[2.1.]{\it \textbf{Normalization (Min-Max):}} Scale input features to the range $\left[0,1\right]$ using min-max normalization to ensure equal feature weighting
				$$
				\tilde{x}_i = \dfrac{x_i - \min(x)}{\max(x)-\min(x)}
				$$
				\item[2.2.]{\it \textbf{Windowing:}} Split data into training and testing data and get a sliding window of values with fixed-length $T$ (shape: $T\times 3$).
				$$
				X_{\text {window }}=\{\tilde{x}_{t-T-1}, \tilde{x}_{t-T}, \ldots, \tilde{x}_{t-1}\} 
				$$
			\end{itemize}
			\item{\it \textbf{3. Encoder (Transformer):}}
			\begin{itemize}
				\item[3.1.]{\it \textbf{Embedding layer:}} Project each $\tilde{x}_{i}$ into high-dimension vectors $d$
				$$
				E_i=W_e \tilde{x}_{i}+b_e,\qquad i = \{t-T-1,\cdots,t-1\}
				$$
				where $W_e\in\mathbb{R}^{d\times 3}$, $b_e\in\mathbb{R}^{d}$.
				\item[3.2.]{\it \textbf{Positional encoding:}} Add time-order information to embeddings (positional encodings) using sine/cosine waves of varying frequencies.
				$$
				\begin{gathered}
					P E_{(\text{pos}, 2 k)}=\sin \left(\frac{\text{pos}}{10000^{2 k / d}}\right), \quad P E_{(\text{pos}, 2 k+1)}=\cos \left(\frac{\text{pos}}{10000^{2 k / d}}\right) \\
					H_i^{(0)}=E_i+P E_i 
				\end{gathered}
				$$
				\item[3.3.]{\it \textbf{Transformer layers:}}
				For each layer $\ell\in\{1,\cdots,L\}$:
				\begin{itemize}	
					\item[3.3.1.]~{\it \textbf{Multi-head self-attention:}} For each head $m$, split $H^{(\ell-1)}$ into $h$ heads 
					$$
					\text { Attention }_m=\operatorname{Softmax}\left(\frac{Q_m K_m^T}{\sqrt{d_k}}\right) V_m, 
					$$
					where
					$$Q_h=H^{(\ell-1)} W_m^Q,K_m=H^{(\ell-1)}W_m^K,V_m=H^{(\ell-1)}W_m^V,d_k=d/h.$$
					\item[3.3.2.]~{\it \textbf{Concatenate heads:}}
					$$
					\operatorname{MultiHead}=\text{Concat}\left[\text { Attention }_1, \cdots, \text { Attention }_h\right] W^O 
					$$
					where $W^O \in\mathbb{R}^{h\cdot d_v\times d}$.
					\item[3.3.3.]~{\it \textbf{Residual connections and LayerNorm::}}
					In deeper layers, each module is wrapped with a residual connection followed by layer normalization:
					$$
					H^{(\ell)}_{\text{attn}}=\operatorname{LayerNorm}(H^{(\ell-1)}+\operatorname{MultiHead})
					$$

					\item[3.3.4.]~{\it \textbf{Feed-Forward Network (FFN):}} Non-linear transformation of features.
					$$
					H^{(\ell)}_{\text{ffn}}=\operatorname{ReLU}\left(H^{(\ell)}_{\text{attn}}W_1+b_1\right) W_2+b_2 
					$$
					where $W_1\in\mathbb{R}^{d\times d_{\text{ffn}}}$ and $W_2\in\mathbb{R}^{d_{\text{ffn}\times d}}$.
					\item[3.3.5.]~{\it \textbf{Second residual connections and LayerNorm::}}
					$$
					H^{(\ell)}=\operatorname{LayerNorm}\left(H^{(\ell)}_{\text{attn}}+H^{(\ell)}_{\text{ffn}}\right)
					$$
				\end{itemize}
			\end{itemize}
			\item{\it \textbf{4. Decoder (GRU with attention):}}
			\begin{itemize}
				\item[4.1.]{\it \textbf{Sequence modeling:}} Process the Transformer’s final-layer output with GRU cells
				$$
				H^{(L)} = \{ H^{(L)}_{t-T-1},\cdots, H^{(L)}_{t-1}\}.
				$$
				The GRU equations:
				$$
				\begin{gathered}
					z_i=\sigma\left(W_z [h_{i-1}, H_{i}^{(L)}]\right) \quad (\text{Update gate})\\
					r_i=\sigma\left(W_r [h_{i-1}, H_{i}^{(L)}]\right)  \quad (\text{Reset gate})\\
					\tilde{h}_i=\tanh \left(W_h [r_i \odot h_{i-1}, H_{i}^{(L)}]\right) \quad (\text{Candidate state})\\
					h_i=\left(1-z_i\right) \odot h_{i-1}+z_i \odot \tilde{h}_i  \quad (\text{Hidden state})
				\end{gathered}
				$$
				\item[4.2.]{\it \textbf{Prediction}}
				Use the final GRU hidden state $h_{t-1}$ to predict the price $\hat{P}_t$:
				$$
				\tilde{P}_t=W_p h_{t-1} +b_p,
				$$
				where $W_p\in\mathbb{R}^{1\times d_{\text{gru}}}$ and $b_p\in\mathbb{R}$.
			\end{itemize}
			\item{\it \textbf{5. Output (post-processing):}} Convert predictions back (denormalization) to original price scale.
			$$
			\hat{P}_{t}^{\text {original}}=\tilde{P}_{t}.\left(P_{\max}-P_{\min}\right)+P_{\min}
			$$	
		\end{description}
	}
	Figure \ref{fig:fig4} illustrates the structure of this model.
	The main idea of our proposed model is to treat the historical cryptocurrency prices, trading volumes, and the Fear and Greed Index as a sequence of tokens, leveraging the self-attention mechanism to capture long-range dependencies while using GRU to detect sequential patterns and short-term fluctuations across different time steps. 
	As shown in Figure \ref{fig:fig4}, the historical data is first encoded as input tokens, then it goes through embedding and positional encoding before entering the Transformer layers. 
	The encoder outputs are fed into the GRU decoder, which applies attention over the encoder's memory to make predictions.
	
	\section{Exploratory Data Analysis}\label{sec:Analysis}
	
	In our analysis, we consider the top two prominent cryptocurrencies with the highest cryptocurrency market capitalization: Bitcoin and Ethereum.
	The daily data was downloaded from the website \url{https://coinmarketcap.com} (last accessed on 4 April 2025), where we consider the daily closing prices and the trading exchange volume in USD for both digital assets. Table \ref{Table:HistoricalData} provides a detailed summary of the respective cryptocurrency price datasets analyzed in this study. The table includes information on the start date, end date, and the total number of records for each dataset, offering a comprehensive overview of the temporal coverage and dataset size for Bitcoin and Ethereum. These details are crucial for understanding the scope and reliability of the data utilized in subsequent analyses.
	\begin{table}[htb!]
		\caption{Historical data of Bitcoin and Ethereum prices.}
		\label{Table:HistoricalData}
		\centering{
			\footnotesize
			\renewcommand{\arraystretch}{1.5}
			\setlength{\tabcolsep}{0.35cm}
			\begin{tabular}{lccc}
				\toprule
				\textbf{Cryptocurrency} & \textbf{Start Date} & \textbf{End Date} & \textbf{Number of Records} \\
				\midrule
				Bitcoin & September 17, 2014& February 28, 2025& 3818\\
				Ethereum & November 9, 2017& February 28, 2025 & 2669\\
				\bottomrule
		\end{tabular}}  
	\end{table}
	Several studies have utilized the fear and greed index (FGI), Google search index (GSI), and Twitter data to explore how sentiment influences cryptocurrency prices \cite{Kristoufek2013, Urquhart2018,Kao2024}. 
	The crypto fear and greed index (FGI) was introduced by \texttt{Alternative.me}, \url{https://alternative.me/crypto/fear-and-greed-index} (last accessed on 4 April 2025), on February 1, 2018, so there is no official FGI data available before this date.
	In order to study the impact of the FGI on cryptocurrency prices prior to February 2018, we propose the following proxy measures using the social media sentiment and Google trends data related to cryptocurrency, specifically for Bitcoin and Ethereum:
	$$
	\text{FGI } = w_1\text{ Social Media Sentiment } + w_2\text{ Google Trends }
	$$
	In our research we assign equal weights ($w_1=w_2=0.5$) to both indicators.
	{\color{black}
		The setting of equal weights was based on a simplifying assumption to ensure balanced contribution from both components, given the absence of a universally accepted weighting scheme in existing literature. 
		While there are studies that have used both social media sentiment and Google Trends as proxies for investor sentiment (e.g., \citet{Mai2018}), specific weights often depend on empirical tuning or are equally weighted when prior knowledge does not favor one source over the other. Our choice aligns with this common practice in early-stage indicator construction, where equal weights are used to preserve interpretability and neutrality (e.g., \citet{Garcia2015}).}
	
	We utilized the Twitter Intelligence Tool (Twint) and Twitter’s API to collect historical social media data from Twitter (now X) before February 1, 2018 using the following keywords with hashtags, \#, and Dollar signs, \$:
	\texttt{Crypto, Cryptocurrency, digital currency, Bitcoin, Ethereum, BTC, ETH}. 
	Then we used the open source \texttt{Python} library Valence Aware Dictionary and Sentiment Reasoner (VADER) to classify the input statement social media sentiment according to a score ranged from $-1$ to 1, where a score from 1 to $-0.05$ stands for Negative sentiment, $-0.05$ to $0.05$ stands for Neutral sentiment, and $0.05$ to 1 stands for Positive sentiment. 
	To collect the historical Google Trends data before February 2018, we enter the aforementioned keywords in the search bar of \url{https://trends.google.com/trends} (last accessed on 4 April 2025) and calculate the average score to get the daily search interest values, which range from 0 to 100.
	
	After that, we calculate the FGI score using the formula:
	\begin{equation}\label{eq:VADER}
		\text{FGI} = \frac{1}{2}\bigg [\frac{\text{Score from VADER} + 1}{2}\times 100 + \text{Score Google Trends}\bigg ]
	\end{equation}
	The FGI ranges from 0 to 100, where $0-24$ represents extreme market fear, $25-49$ means market fear, $50-74$ indicates market greed, and $75-100$ represents extreme market greed.
	
\begin{figure}[htb!]
		\includegraphics[width=1\textwidth,height=0.23\textheight]{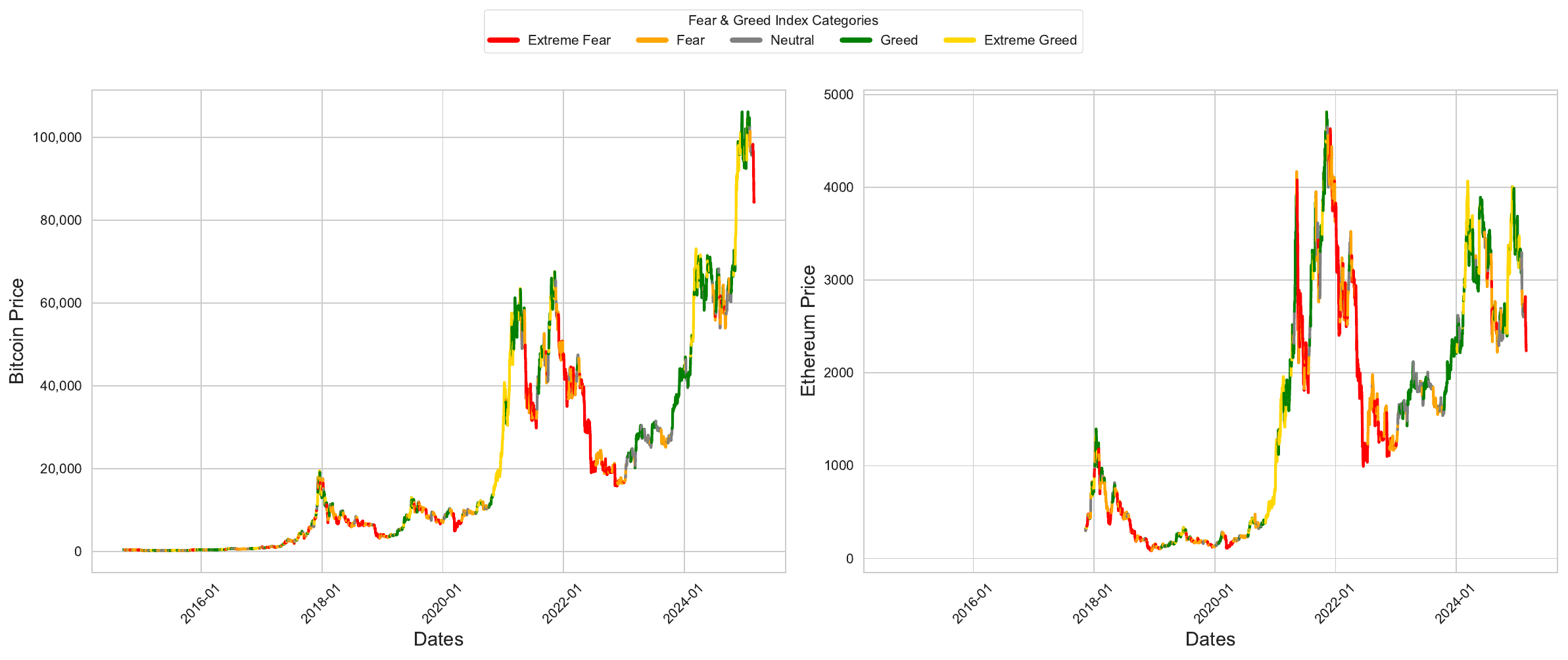}
		\vspace{-0.1cm}
		\caption{Bitcoin daily prices (left) and Ethereum prices (right), with trends color-coded according to fear and greed index (FGI) categories.}
		\label{fig:fig5}
	\end{figure}

\begin{figure}[htb!]
		\includegraphics[width=1\textwidth,height=0.2\textheight]{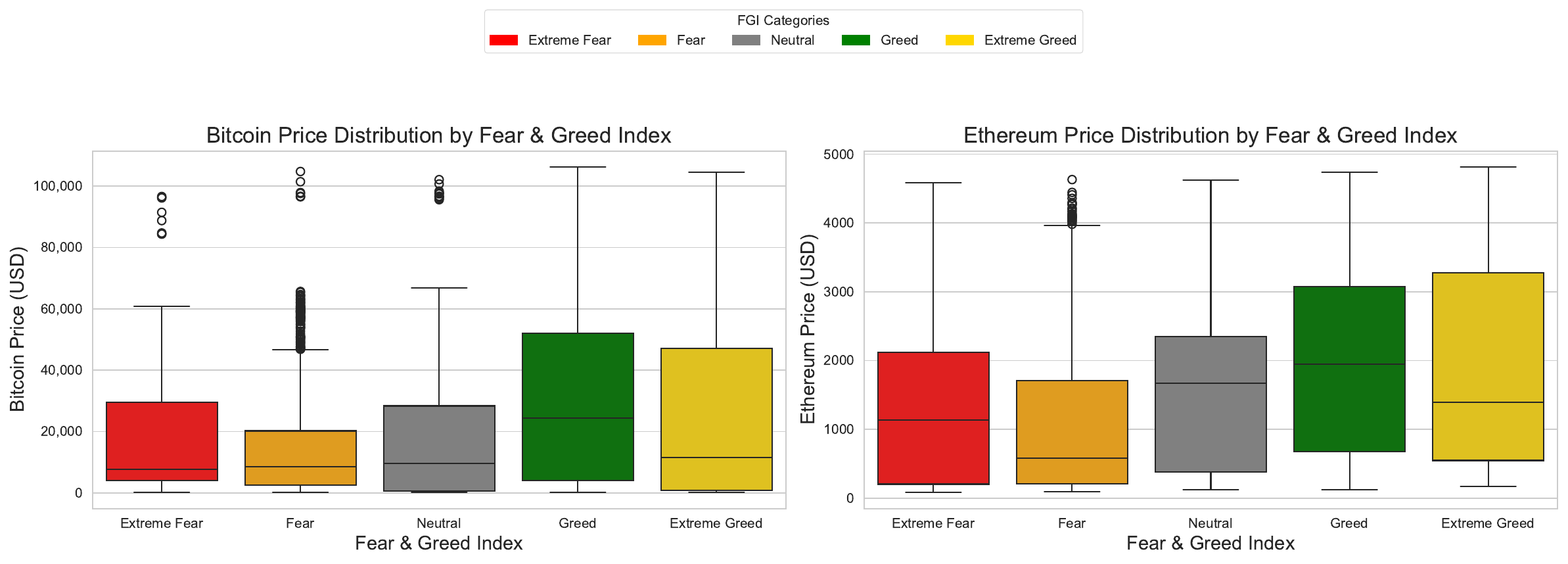}
		\vspace{-0.1cm}
		\caption{Box-plots of Bitcoin daily prices (left) and Ethereum prices (right) categorized by the fear and greed index (FGI).}
		\label{fig:fig6}
	\end{figure}

	Figures \ref{fig:fig5} shows the daily prices of Bitcoin from September 17, 2014 to February 28, 2025 and Ethereum from November 9, 2017 to February 28, 2025.
	It can be seen that the prices of both currencies have increased exponentially over time, showing a strong cyclic pattern in relation to FGI.
	Although the long-term trend remains upward, the prices of Bitcoin and Ethereum often decline during fear/extreme fear periods followed by recoveries during greed/extreme greed times.
	The boxplots in Figure \ref{fig:fig6} show the distribution of cryptocurrency prices for categories of FGI sentiments.
	When fear sentiment dominates, prices tend to cluster at lower values (right-skewed), suggesting most trades happen at depressed levels. 
	Extreme fear shows a more balanced distribution (symmetric) but with slightly higher prices, hinting at potential stabilization or cautious buying. Neutral sentiment shifts prices upward (left-skewed), reflecting modest optimism. 
	Greed pushes prices higher (left-skewed), with more trades concentrated at premium values. 
	Finally, extreme greed results in a right-skewed pattern indicating speculative spikes and increase volatility, likely from FOMO-driven buying. 
	
\begin{figure}[htb!]
			\centering
			\includegraphics[width=1\textwidth,height=0.41\textheight]{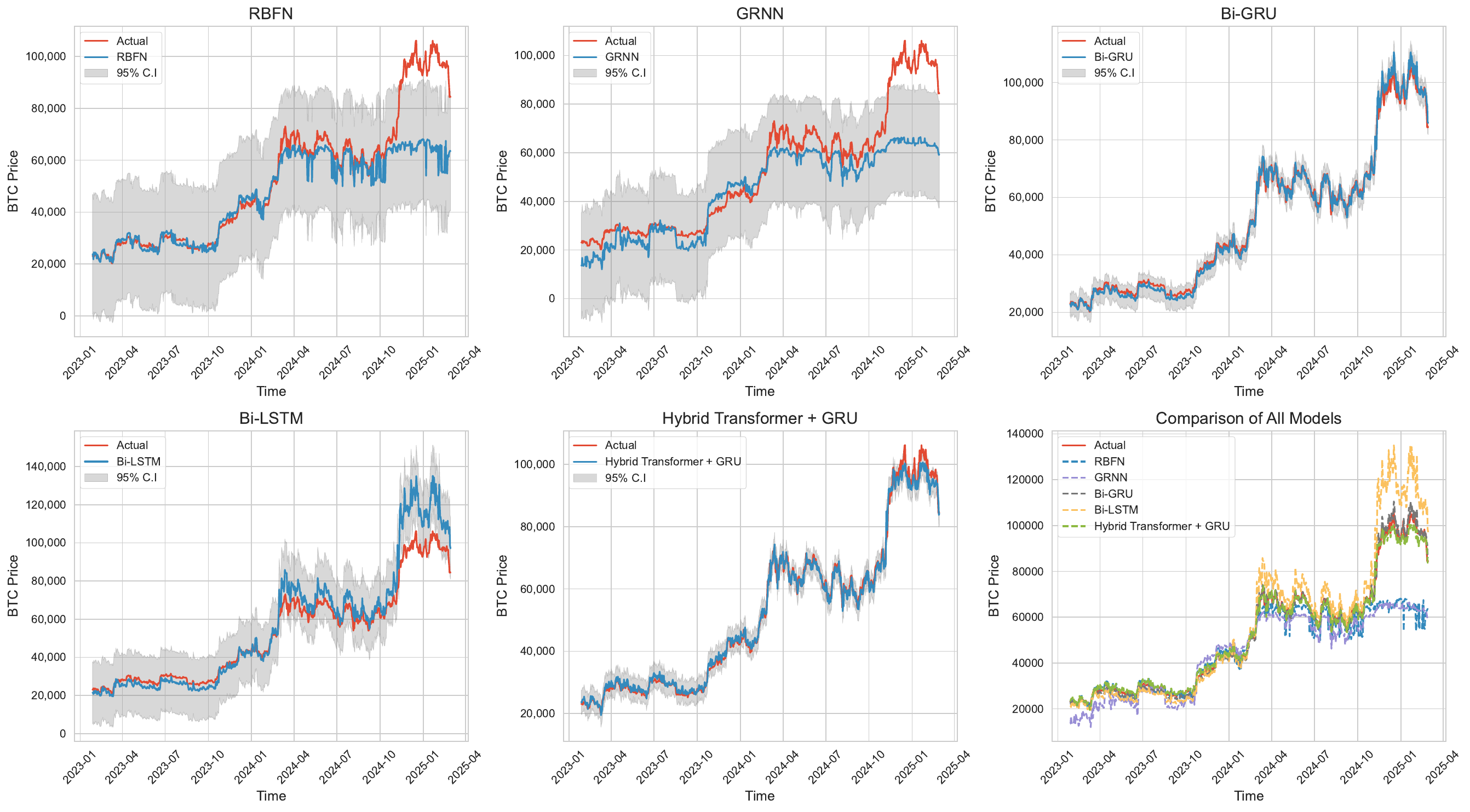}
			\vspace{-0.3cm}
			\caption{Comparison of the proposed hybrid Transformer-GRU model with four competing deep learning networks for Bitcoin price prediction.}
			\label{fig:fig7}
		\end{figure}

\begin{figure}[htb!]
		\centering
		\includegraphics[width=1\textwidth,height=0.41\textheight]{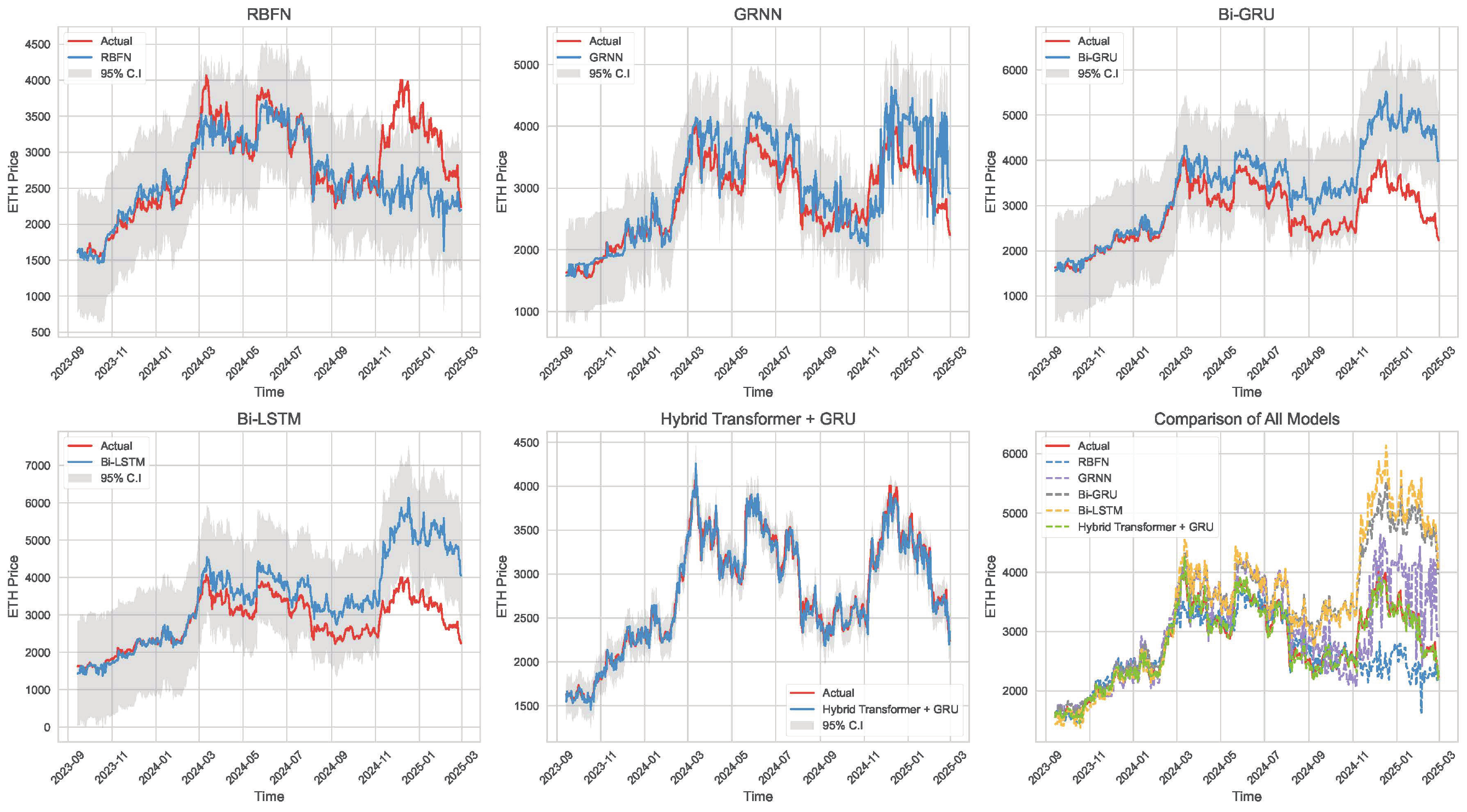}
		\vspace{-0.3cm}
		\caption{Comparison of the proposed hybrid Transformer-GRU model with four competing deep learning networks for Ethereum price prediction.}
		\label{fig:fig8}
	\end{figure}

	{\color{black}We began our analysis by splitting the dataset into an 80-20 train-test split.
		For Bitcoin analysis, the training set covers the period from September 17, 2014, to January 26, 2023, while the testing set runs from January 27, 2023, to February 28, 2025.
		In the case of Ethereum, the training data spans from November 9, 2017, to September 13, 2023, and the testing period extends from September 14, 2023, to February 28, 2025.
		Then, we normalize both datasets using the Min-Max normalizing.
		It is worth to note that we prefer using the min-max normalization over z-score standardization because the cryptocurrency prices tend to be non-stationary and extremely volatile. 
		In such cases, the min-max normalization is more robust compared to z-score standardization.
		After that, we fit the model \ref{eq:model} using the five neural networks (RBFN, GRNN, BiGRU, BiLSTM, Hybrid transformer $+$ GRU) on the training data and predict the prices of Bitcoin and Ethereum using the testing data.} 
	Figures \ref{fig:fig7} and \ref{fig:fig8} show the prediction results along with their corresponding 95\% prediction interval compared with the actual prices of Bitcoin and Ethereum, respectively. 
	As evident from these figures, we observe that our proposed model outperforms the other approaches in capturing price movements.
	
	\clearpage
		
	
	In order to measure the prediction accuracy of the five models, we calculate the mean squared error (MSE), root mean squared error (RMSE), mean absolute error (MAE), and mean absolute percentage error (MAPE) using the testing data.
	The smaller the value the metric, the better the prediction neural network model.
	These metrics can be expressed mathematically as follows:
	\begin{align}
		\text{MSE} &= \frac{1}{n}\sum_{t=1}^{n}(y_t- \hat{y}_t) ^2\\
		\text{RMSE} &= \sqrt{\frac{1}{n}\sum_{t=1}^{n}(y_t- \hat{y}_t) ^2}\\
		\text{MAE} &= \frac{1}{n}\sum_{t=1}^{n}(\mid y_t- \hat{y}_t\mid )\\
		\text{MAPE} &= \frac{1}{n}\sum_{t=1}^{n}\left (\frac{\mid y_t- \hat{y}_t\mid}{y_t} \right)\times 100\%
	\end{align}
	where $y_t$ and $\hat{y}_t$ are the actual price and predicted price of cryptocurrency at time $t$.
	{\color{black}
		We also apply Friedman statistic and post-hoc analysis to compare between the performance of the five models.
		The Friedman test statistic defined by 
		\begin{equation}\label{eq:Fridman}
			\chi_F^2=\frac{12 n}{k(k+1)}\sum_{j=1}^k \bar{R}_j^2-3n(k+1),
		\end{equation}
		where $n$ is the number of datasets (blocks), $k =5$ is the number of models, and $\bar{R}_j$ is the average rank of the $j$-th treatment across all blocks is used to test the hypotheses:
		\begin{align*}
			H_0 &:~\text{All models have identical performance}.\\
			H_a &:~\text{At least one model performs significantly differently from the others}.
		\end{align*}
		If Friedman test rejects $H_0$ (p-value is less than the significant level $\alpha$, using chi-square distribution with $k-1$ degrees of freedom), one can use post-hoc tests to identify which model pairs differ. 
		In this article, we consider using the Wilcoxon Signed-Rank test (with Bonferroni correction using a level of significance $\alpha^\star = \alpha/5$):
		\begin{equation}\label{eq:Wilcoxon}
			R = \min(R^{+}, R^{-}),
		\end{equation}
		where 
		$R^{+} =$ sum of positive ranks and $R^{-} =$ sum of negative ranks of difference in errors.
	}
	
	Tables \ref{Table:CompararBTC}$-$\ref{Table:CompararETH} present the comparison of performance metrics results for the proposed hybrid transformer $+$ GRU network and the other four neural networks for predicting the prices of Bitcoin and Ethereum respectively.
	From these two tables, it is seen that the proposed model substantially superior its compositors achieving more precise predictions as it always gives small predictions errors.
	For example, as shown in Table \ref{Table:CompararBTC}, when comparing the four compositors machine learning models, the BiGRU network achieves the smallest MSE value of $4,358,457$ for Bitcoin price prediction. 
	Whereas, the MSE value for the proposed hybrid transformer $+$ GRU model, $3,818,128$.
	For Ethereum price prediction, as shown in Table \ref{Table:CompararETH}, the proposed model demonstrates excellent performance with an MSE of $11,344.686$.
	This is approximately 18 times smaller than the RBNF model's MSE ($203,541.169$), 17 times smaller than GRNN's ($194,985.943$), 60 times smaller than BiGRU's ($687,799.984$), and 80 times smaller than BiLSTM's ($907,844.180$).
	From the results in both tables, we can see that the feedforward neural networks (RBFN and GRNN) struggled to accurately predict Bitcoin prices.
	On the other hand, both bidirectional models (BiGRU and BiLSTM) didn't perform as well when it came to predicting Ethereum prices.
	
	\setcounter{table}{1}
	
	\begin{table}[htb!]
		\caption{Performance metrics (MSE, RMSE, MAE, and MAPE) for predicting the prices of Bitcoin using different models.}
		\label{Table:CompararBTC}
		\centering{
			\renewcommand{\arraystretch}{1.3}
			\setlength{\tabcolsep}{0.61cm}
			\begin{tabular}{lcccc}
				\toprule
				\textbf{Model} & \textbf{MSE} & \textbf{RMSE} & \textbf{MAE} & \textbf{MAPE} \\
				\midrule
				RBFN                      & 1.731258e+08 & 13157.727 & 6928.640 &  9.479{\color{black}\%}\\
				GRNN                      & 1.875502e+08 & 13694.897 & 9179.342 & 15.857{\color{black}\%}\\
				BiGRU                    & 4.358457e+06 &  2087.692 & 1559.954 & 3.271{\color{black}\%}\\
				BiLSTM                   & 8.184621e+07 &  9046.889 & 5877.042 & 9.600{\color{black}\%}\\
				Hybrid Transformer + GRU  & 3.818128e+06 &  1954.003 & 1419.972 & 2.825{\color{black}\%}\\
				\bottomrule
		\end{tabular}}  
	\end{table}
	
	\begin{table}[htb!]
		\caption{Performance metrics (MSE, RMSE, MAE, and MAPE) for predicting the prices of Ethereum using different models.}
		\label{Table:CompararETH}
		\centering{
			\renewcommand{\arraystretch}{1.3}
			\setlength{\tabcolsep}{0.69cm}
			\begin{tabular}{lcccc}
				\toprule
				\textbf{Model} & \textbf{MSE} & \textbf{RMSE} & \textbf{MAE} & \textbf{MAPE} \\
				\midrule
				RBFN                      & 203541.169 & 451.155 & 288.453 &  9.362{\color{black}\%}\\
				GRNN                      & 194985.943 & 441.572 & 345.963 & 11.901{\color{black}\%}\\
				BiGRU                    & 687799.984 & 829.337 & 608.416 & 20.735{\color{black}\%}\\
				BiLSTM                   & 907844.180 & 952.809 & 675.427 & 22.640{\color{black}\%}\\
				Hybrid Transformer + GRU  & 11344.686  & 106.511 &  78.809 &  2.755{\color{black}\%}\\
				\bottomrule
		\end{tabular}}  
	\end{table}

	\begin{table}[htb!]
		\color{black}\caption{Pairwise model comparisons for Bitcoin price prediction using Wilcoxon signed-rank test with Bonferroni correction}
		\label{Table:WilcoxonBTC}
		\centering{
			\renewcommand{\arraystretch}{1.1}
			\setlength{\tabcolsep}{0.28cm}
			\begin{tabular}{lllllc}
				\toprule
				\textbf{Model 1} & \textbf{Model 2} & \textbf{Wilcoxon} & \textbf{Raw} & \textbf{Bonferroni } & \textbf{Significant}\\
				&  & \textbf{test} & \textbf{p-value} & \textbf{corrected p-value} & \textbf{$^\star$}\\
				\midrule
				RBFN&	GRNN&	38606&	1.78E-69&	1.78E-68&	TRUE\\
				RBFN&	BiGRU&	47196&	4.23E-59&	4.23E-58&	TRUE\\
				RBFN&	BiLSTM&	132787&	0.02894864&	0.2894864&	FALSE\\
				RBFN&	Hybrid Transformer + GRU&	34773&	2.20E-74&	2.20E-73&	TRUE\\
				GRNN&	BiGRU&	5058&	3.17E-118&	3.17E-117&	TRUE\\
				GRNN&	BiLSTM&	47851&	2.41E-58&	2.41E-57&	TRUE\\
				GRNN&	Hybrid Transformer + GRU&	2604&	2.64E-122&	2.64E-121&	TRUE\\
				BiGRU&	BiLSTM&	20834&	1.14E-93&	1.14E-92&	TRUE\\
				BiGRU&	Hybrid Transformer + GRU&	128651&	0.004209894&	0.042098938&	FALSE\\
				BiLSTM&	Hybrid Transformer + GRU&	21898&	4.06E-92&	4.06E-91&	TRUE\\
				\bottomrule
		\end{tabular}}  
	\end{table}

	\begin{table}[htb!]
		\color{black} \caption{Pairwise model comparisons for Ethereum price prediction using Wilcoxon signed-rank test with Bonferroni correction}
		\label{Table:WilcoxonETH}
		\centering{
			\renewcommand{\arraystretch}{1.1}
			\setlength{\tabcolsep}{0.28cm}
			\begin{tabular}{lllllc}
				\toprule
				\textbf{Model 1} & \textbf{Model 2} & \textbf{Wilcoxon} & \textbf{Raw} & \textbf{Bonferroni } & \textbf{Significant}\\
				&  & \textbf{test} & \textbf{p-value} & \textbf{corrected p-value} & \textbf{$^\star$}\\
				\midrule
				RBFN&	GRNN&	47616&	2.49E-11&	2.49E-10&	TRUE\\
				RBFN&	BiGRU&	20411&	2.19E-46&	2.19E-45&	TRUE\\
				RBFN&	BiLSTM&	15193&	5.61E-56&	5.61E-55&	TRUE\\
				RBFN&	Hybrid Transformer + GRU&	17810&	4.73E-51&	4.73E-50&	TRUE\\
				GRNN&	BiGRU&	29575&	8.83E-32&	8.83E-31&	TRUE\\
				GRNN&	BiLSTM&	26767&	5.93E-36&	5.93E-35&	TRUE\\
				GRNN&	Hybrid Transformer + GRU&	5615&	5.43E-76&	5.43E-75&	TRUE\\
				BiGRU&	BiLSTM&	46690&	4.11E-12&	4.11E-11&	TRUE\\
				BiGRU&	Hybrid Transformer + GRU&	4900&	1.31E-77&	1.31E-76&	TRUE\\
				BiLSTM&	Hybrid Transformer + GRU&	4839&	9.48E-78&	9.48E-77&	TRUE\\
				\bottomrule
		\end{tabular}}  
	\end{table}
	
	The Friedman test results show a statistically significant difference in performance among the five Bitcoin price prediction models $(\chi^2 = 1419.34,~\text{p-value} < 0.001)$, indicating that the models did not perform equally.
	As shown in Table \ref{Table:WilcoxonBTC}, post hoc pairwise comparisons using the Wilcoxon signed-rank test with Bonferroni correction revealed that the Hybrid Transformer + GRU model significantly outperformed RBFN, GRNN, and BiLSTM. 
	However, its performance was not significantly better than that of the BiGRU model $(\text{p-value} = 0.0421 > \alpha^\star= 0.005)$. 
	Additionally, BiGRU showed significantly stronger performance compared to both Bi-LSTM and GRNN, while RBFN was the weakest model overall, outperformed by all others except BiLSTM. 
	Overall, both the BiGRU and the Hybrid Transformer + GRU models performed the best for Bitcoin price prediction, with only slight variations between them.
	
	Similar to the Bitcoin results, the Friedman test shows significant performance differences among the five Ethereum prediction models $(\chi^2 = 747.76,~\text{p-value} < 0.001)$. 
	As shown in Table \ref{Table:WilcoxonETH}, post hoc Wilcoxon signed-rank tests with Bonferroni correction showed that the Hybrid Transformer + GRU model significantly outperformed all other models, including BiGRU and BiLSTM. 
	The results are consistent with the Bitcoin analysis, confirming the robustness of our proposed model across different digital cryptocurrencies.
	
	\section{Conclusions and Recommendation}\label{sec:Conclusions}
	
	The results obtained in this study show better performance of a novel attention-based hybrid Transformer model in predicting cryptocurrency prices compared with two feedforward neural networks (radial basis function, RBFN, and general regression neural network, GRNN) and two deep learning models (bidirectional gated recurrent unit, BiGRU, and bidirectional long-short-term memory, BiLSTM).
	Our proposed model builds on the same Transformer architecture, the foundation behind AI systems like ChatGPT and DeepSeek.
	{\color{black}The academic contribution of this study lies in demonstrating how a hybrid architecture that combines long-range pattern recognition with short-term temporal modeling can enhance time series forecasting. 
		Practically, our model offers potential for improving real-time decision-making in cryptocurrency markets, particularly for traders and analysts seeking data-driven strategies based on historical patterns.}
	We conducted the comparison study using the two leading digital assets, Bitcoin and Ethereum based on historical data with a one-time-step lag.
	When predicting Bitcoin prices, the proposed model achieves $6-7$ times lower RMSE, $4-6$ times lower MAE, and $3-6$ times lower MAPE compared to the RBFN and GRNN models.
	For Ethereum, our model reduces RMSE by about $4$ times, MAE by $3-4$ times, and MAPE by $3-4$ times compared to RBFN and GRNN.
	On the other hand, when we predict the Bitcoin prices, the proposed model achieves $1-5$ times lower RMSE, $1-4$ times lower MAE, and $1-3$ times lower MAPE compared to the BiGRU and BiLSTM models.
	Moreover for Ethereum, our model reduces RMSE, MAE, and MAPE by approximately $8-9$ times compared to BiGRU and BiLSTM models.
	This study has certain limitations. First, it focuses only on Bitcoin and Ethereum, which, while dominant, do not represent the full spectrum of cryptocurrency behavior. 
	Second, the model uses a fixed one-time-step lag, which may not capture more complex temporal dependencies. 
	Third, we relied primarily on historical price, volume, and the Fear and Greed Index; other macroeconomic or blockchain-based indicators could further enrich the model.
	Future research could address these limitations by applying the model to a wider range of univariate and multivariate digital assets, as well as stock markets (e.g., as explored in \citet{Yan2024}), testing different lag structures, and exploring alternative hybrid architectures such as Transformer + LSTM or Transformer + BiLSTM/BiGRU.

\end{document}